\pdfoutput=1

\documentclass[11pt]{article}

\usepackage{soul}
\usepackage{tcolorbox}
\tcbuselibrary{listings, skins, breakable}
\newtcolorbox{highlightedblock}{
  enhanced,
  breakable,
  colback=yellow!20,
  colframe=yellow!20,
  boxrule=0pt,
  sharp corners,
  top=2pt,
  bottom=2pt,
  left=4pt,
  right=4pt,
}

\usepackage[final]{acl}
\usepackage{subcaption}
\usepackage{soul}  
\usepackage{times}
\usepackage{latexsym}
\usepackage{amsthm}
\theoremstyle{definition}
\newtheorem{definition}{Definition}
\usepackage[T1]{fontenc}

\usepackage[utf8]{inputenc}

\usepackage{microtype}

\usepackage{inconsolata}

\usepackage{float}
\usepackage{svg}
\usepackage{hyperref}
\usepackage{graphicx}
\usepackage{url}
\usepackage{booktabs}
\usepackage{multirow}
\usepackage{lineno}
\usepackage{soul}
\usepackage{amsmath}
\usepackage{algorithm}
\usepackage{algpseudocode}
\usepackage{verbatim}
\usepackage{float}
\usepackage{amsfonts}

\title{Quantifying and Mitigating Selection Bias in LLMs: A Transferable LoRA Fine-Tuning and Efficient Majority Voting Approach}

\author{
Blessed Guda\textsuperscript{1,2},
Lawrence Francis\textsuperscript{1},
Gabrial Zencha Ashungafac\textsuperscript{1},\\
\textbf{Carlee Joe-Wong\textsuperscript{1,2},
Moise Busogi\textsuperscript{1,2}}\\[3pt]
\textsuperscript{1}Carnegie Mellon University Africa, Kigali, Rwanda\\
\textsuperscript{2}Carnegie Mellon University, Pittsburgh, PA 15213, USA\\[3pt]
\texttt{\{blessedg, lfrancis, gzenchaa, cjoewong, mbusogi\}@andrew.cmu.edu}
}

\begin{document}
\maketitle
\begin{abstract}
Multiple Choice Question (MCQ) answering is a widely used method for evaluating the performance of Large Language Models (LLMs). However, LLMs often exhibit selection bias in MCQ tasks, where their choices are influenced by factors like answer position or option symbols rather than the content. This bias undermines the reliability of MCQ as an evaluation framework. Most existing selection bias metrics require answer labels and measure divergences between prediction and answer distributions, but do not fully capture the consistency of a model’s predictions across different orderings of answer choices. Existing selection bias mitigation strategies have notable limitations: majority voting, though effective, is computationally prohibitive; calibration-based methods require validation sets and often fail to generalize across datasets.
To address these gaps, we propose three key contributions: (1) a new unsupervised label-free \textbf{Permutation Bias Metric (PBM)} that directly quantifies inconsistencies in model predictions across answer permutations, providing a more precise measure of selection bias, (2) an efficient majority voting approach called Batch Question-Context KV caching (BaQCKV), to significantly reduce computational costs while preserving bias mitigation effectiveness, and (3) an unsupervised Low-Rank Adaptation (LoRA)-1 fine-tuning strategy based on our proposed metric and the BaQCKV that mitigates selection bias, providing a computationally efficient alternative that maintains model generalizability. Experiments across multiple MCQ benchmarks demonstrate that our approaches reduce bias, increasing consistency in accuracy while minimizing computational costs.
\end{abstract}

\section{Introduction}
Selection bias in Large Language Models (LLMs) has been increasingly recognized as a significant challenge, particularly in multiple-choice question (MCQ) answering tasks~\citep{wei2024unveiling,zheng2024large, zong2023fool}. This bias occurs when models exhibit a preference for certain answer choices based on factors like their position or symbolic representation, rather than the option content \citep{wei2024unveiling}. For instance, LLMs may disproportionately favor the last option or option “A” across different questions. Such biases are especially problematic in evaluation settings, where multiple-choice formats are widely used for example, in standardized testing, professional certification exams, and educational assessments. These biases undermine the fairness and reliability of model evaluations, as they can lead to inconsistent answers across equivalent permutations, eroding trust in LLM-based decision systems. 

The presence of selection bias in LLMs was highlighted by \cite{zheng2024large}, demonstrating how factors like answer position and symbolic representation can lead to systematic errors in MCQ answering.
Effectively addressing selection bias requires a well-defined metric for bias quantification.
Several metrics have been proposed to measure the selection bias such as the Choice Kullback-Leibler Divergence (CKLD) \citep{choi2024mitigatingselectionbiasnode},  Standard Deviation of Recalls (RStd) \citep{zheng2024large}, and Relative Standard Deviation (RSD) \citep{croce2021robustbench, reif2024beyond}, which primarily evaluate bias in terms of divergence from ground truth distributions (i.e., CKLD) or variability in class-wise performance (i.e., RStd and RSD). However, they do not adequately capture the bias exhibited by models to option permutations. Also, the Fluctuation Rate proposed by \citep{wei-etal-2024-unveiling} only considers two permutations of the options, which may not capture the full permutation bias.   
We therefore introduce a new permutation bias  metric (PBM) that evaluates selection bias in LLM without requiring ground truth distributions while considering all possible option permutations. The primary intuition behind our metric is that \textit{logically, an answer's correctness does not change based on its position in a list of options and we therefore want language models to possess this behaviour}. 

Addressing the problem of bias requires not just quantifying but also mitigating bias. Prior mitigation strategies like majority voting \cite{zong2023fool} that aggregates predictions across all permutations of answer choices - has been shown to reduce bias. However, its computational cost increases factorially with the number of choices, making it impractical for real-time inference. Thus, a key challenge is to develop an efficient method for bias quantification and bias mitigation that can be integrated into real-world systems. We therefore propose \textbf{Batch Question-Context KV caching} (BaQCKV), an efficient implementation of majority voting that reduces computational cost considering all permutations. Additionally, we introduce an unsupervised Low-Rank Adaptation \cite{hu2021loralowrankadaptationlarge} finetuning strategy that optimizes the model on our proposed metric.
\\
Our contributions can be summarized as follows:
\begin{itemize}
    \item We propose a novel, unsupervised, and label-free Permutation Bias metric (PBM) that captures inconsistencies in model predictions across all permutations of answer choices. Unlike prior metrics, it requires no access to ground-truth labels and directly measures permutation sensitivity.
    \item We introduce \textbf{BaQCKV (Batch Question-Context KV caching)}, a computationally efficient variant of majority voting that significantly reduces the overhead associated with evaluating all permutations of multiple-choice questions.
    \item We develop a LoRA-based fine-tuning strategy that leverages our proposed bias metric as a differentiable objective, enabling parameter-efficient debiasing without the need for labeled data or full model retraining. 
 
\end{itemize}
Our efficient BaQCKV method achieves token savings of up to \textbf{54.4\%}, while our lightweight unsupervised LoRA-1 (LoRA Rank = 1) fine-tuning reduces the PBM bias by an average of \textbf{58\%} and improves standard deviation of accuracy by \textbf{27\%}, outperforming existing approaches. 
BaQCKV is particularly well-suited for evaluation or deployment scenarios where deterministic and fully permutation-invariant responses are required, as it can achieve 0 bias but with additional compute and latency. In contrast, LoRA-1 fine-tuning offers a lightweight, one-pass inference alternative for practical large-scale LLM deployments or latency-sensitive settings. Together, these contributions  can lead to a unified framework for quantifying and mitigating selection bias in LLMs, particularly in the context of multiple-choice question answering.
 


\section{Related Work}\label{sec:rel_works}

Large Language Models (LLMs) exhibit systematic selection biases in multiple-choice question answering (MCQA), favoring options by position (e.g., last choice) or by identifier (e.g., option “A”) rather than semantic content. 
In this review, we focus on existing approaches for bias quantification and mitigation.

\subsection{Bias Evaluation in LLMs}\label{lit:evaluate}

Bias quantification typically measures divergence between predicted and ground-truth answer distributions. Choice Kullback–Leibler Divergence (CKLD) \cite{choi2024mitigatingselectionbiasnode}  measures the KL divergence between the model’s predicted answer frequency and the ground-truth (i.e, how often the correct answer is A, B., etc.). Other metrics focus on variability in per-option accuracy and recall. The Relative Standard Deviation (RSD)  \cite{reif2024beyond} and  the Standard Deviation of Recall (RStd) \cite{zheng2024large} assess variability in per-option accuracy or recall, respectively, revealing positional preference, but ignore how predictions change under option reordering. While useful, such label-dependent metrics fail to capture \textit{inconsistency across permutations} They do not consider if a model would answer the same question differently when choices are presented in a different option permutation, since they only evaluate against the single correct label in the original ordering.

To capture such inconsistency, \citet{wei2024unveiling}\ introduced Fluctuation Rate (FR), a label-free metric that measures answer changes when options are reversed. While FR highlights instability, it is limited to two permutations and only detects discrete flips, ignoring confidence shifts and being non-differentiable, which limits its use in fine-tuning.

In summary, current metrics are either label-dependent (CKLD, RSD/RStd) or permutation-limited (FR), offering incomplete views of bias. To address this, we propose a \textbf{ \textit{permutation-sensitive, label-agnostic metric} }that captures prediction consistency across all answer orderings. Our approach enables broad applicability on unlabeled datasets and introduces a \textbf{\textit{differentiable objective for debiasing during fine-tuning}} (section \ref{sec:LoRA1-1}).

\subsection{Bias Mitigation Strategies}\label{sec:mitigate}
Several research works have explored various strategies to mitigate selection bias in LLMs, ranging from calibration, voting, to finetuning-based approaches.

\textbf{Calibration-based} methods aim to adjust the model’s output probabilities to compensate for the skewed bias distributions. Most of these methods target recalibration to improve accuracy and not directly to reduce bias. CalibraEval \cite{calibraeval2024} reweights predictions to reduce positional bias during evaluation, while label bias calibration \cite{reif2024beyond} improves accuracy using known statistics. PriDe  \cite{zheng2024large} estimates prior probabilities over option IDs to normalize predictions, effectively reducing RStd. These methods, however, typically need labeled data and apply the same calibration across permutations, thus failing to ensure consistency under reordering. Prompt-based fixes like Auxiliary Option Injection (AOI)  \cite{choi2024mitigatingselectionbiasnode} are simple and model-agnostic but offer limited and prompt-sensitive improvements.

\textbf{Majority voting} aggregates predictions over option permutations to reduce bias \cite{zong2023fool, wei-etal-2024-unveiling}. Though effective, it scales poorly to $k!$ permutations for $k$ options making it impractical. Efficient variants include batch prompting \cite{batchcalibration2024} and random subset voting \cite{guda2024qmos}. Self-consistency \cite{selfconsistency2024, universalsc2024} improves stability by sampling varied reasoning paths, akin to voting. However, all such methods incur significant inference-time cost. Our BaQCKV approach reduces this by reusing computation across permutations.

Instead of repeatedly querying a biased model at inference time, another strategy is to adjust the parameters that induce bias within the model itself. Teacher–student distillation \cite{liusie2024teacherstudenttrainingdebiasinggeneral} transfers debiased behavior from a teacher model into compact models. Bias Node Pruning (BNP) \cite{choi2024mitigatingselectionbiasnode} identifies a bias vector in the final decoder layer and prunes parameters in the LLM's final linear head projection matrix based on their interactions with this vector. While such methods reduce metrics like FR, they may negatively impact the model's performance on other tasks due to irreversible weight pruning or overfitting to specific bias patterns.

In summary, training-time debiasing techniques including knowledge distillation, fine-tuning, and structural pruning aim to internalize bias mitigation and reduce the need for repeated inference-time interventions. However, these often require labeled datasets and can compromise generalization. In contrast, our approach introduces a fully differentiable, label-free bias objective that enables debias fine-tuning to reduce permutation sensitivity. This allows for unsupervised debiasing that generalizes across datasets.


\section{Methodology}\label{methodology}
In this section, we define our bias quantification metric, which accounts for all option permutations and also describe the mitigation strategies - BaQCKV and LoRA-1 fine-tuning.


\subsection{Permutation Bias Metric (PBM)}\label{sec:select_bias_qant}
PBM is based on the intuition that, logically, a model's confidence in an option should be invariant to the permutation of the options. Also, we argue that this quantification should be label-free because the confidence for each option content across all permutations should be constant regardless of option correctness. Let \( Q \) represent a question, and \( O = \{o_1, o_2, \dots, o_n\} \) represent a set of \( n \)  options for the question. A model processes the sequence \( S_{\pi} = Q \oplus O_{\pi} \) for a permutation \( \pi \) of the options \( O \), where \( \oplus \) denotes the concatenation operator. Passing a permutation of the options $S_{\pi}$ through the model assigns probabilities \( P(o_{\pi(i)} \mid Q, O_{\pi}) \) to each option content \( o_i \). Similarly, for a different permutation $\pi^{'}$ it assigns \( P(o_{\pi^{'}(i)} \mid Q, O_{\pi^{'}}) \). We define our selection bias metric mathematically  for a model in Equation (\ref{eqn:our_bias}) as the variance of the probabilities for each option content across all permutations. This will capture how much the model’s confidence fluctuates due to reordering. 
\begin{definition}[Permutation Bias Metric – PBM]\label{def:bias_metric}
Given a question \( Q \) and a set of answer options \( O = \{o_1, o_2, \dots, o_n\} \), the selection bias \( B(Q, O) \) is defined as:
\begin{equation}
    B(Q, O) = \frac{1}{n} \sum_{i=1}^n \text{Var}_{\pi}\left(P(o_{\pi(i)} \mid Q, O_{\pi})\right),
    \label{eqn:our_bias}
\end{equation}
We refer to this selection bias score as the \textbf{Permutation Bias Metric (PBM)},
where:
\begin{equation}
\begin{split}
    \text{Var}_{\pi}\left(P(o_{\pi(i)} \mid Q, O_{\pi})\right) &= \\
    \frac{1}{n!} \sum_{\pi} \biggl( P(o_{\pi(i)} \mid Q, O_{\pi}) & \\
    \quad - \mathbb{E}_{\pi}\left[P(o_{\pi(i)} \mid Q, O_{\pi})\right] \biggr)^2,
\end{split}
\end{equation}
and the expectation over all permutations \( \pi \) is:
\[
\mathbb{E}_{\pi}[P(o_{\pi(i)} \mid Q, O_{\pi})] = \frac{1}{n!} \sum_{\pi} P(o_{\pi(i)} \mid Q, O_{\pi}).
\]
\end{definition}
By defining the \textbf{PBM} to be proportional to the variance of a model's prediction across all permutations, the metric captures the inconsistency in the predictions. A lower variance indicates more stable and consistent predictions across permutations. Therefore, lower values of the \textbf{PBM} are preferred.
\textbf{PBM} is label-free compared to previously proposed metrics like RStd (equation (\ref{eqn::rstd})), FR (equation (\ref{eqn::fr})) and CKLD (equation (\ref{eqn:ckld})) which capture some form of divergence from the answer distribution, requiring the ground truth answers. The RStd and CKLD do not capture any information about the option permutations because they only use the original permutations of the options. This assumes the original permutation corresponds to the fixed answer order provided in the dataset, which is often arbitrary (i.e, randomly assigned during test construction). Also, while \textbf{PBM} goes into the granular confidence level by considering option probabilities, the FR does not capture granular information of the changes in confidence but only checks if the chosen answer (argmax) is the same in the forward and reverse permutations. By considering only 2 permutations, it does not capture information from other permutations. The Fluctuation Rate (FR) is expressed as:

\begin{multline}
    FR = \frac{1}{N} \sum^{N} \Bigl( \operatorname*{argmax}_{i} \bigl( P(O \mid Q, O_{\pi} ) \bigr) \\
    \neq \operatorname*{argmax}_{i} \bigl( P(O \mid Q, reverse(O_{\pi} )) \bigr) \Bigr)
    \label{eqn::fr}
\end{multline}
where the $reverse(O_{\pi})$ function is the reverse permutation of ${O_{\pi}}$ and  $N$ is the total number of questions.

RStd (\(R_{\sigma}\)) is expressed as:

\begin{equation}
    R_{\sigma} = \sqrt{\frac{1}{n} \sum_{i=1}^{n} (r_i - \mu_r)^2} 
    \label{eqn::rstd}
\end{equation}
where \( n \) represents the number of option classes, \( r_i \) is the recall for the \( i \)-th option symbol (e.g. recall for option A, B....), and \( \mu_r \) denotes the average recall across all option symbols. 
\begin{equation}
    CKLD = \sum_{i}^{n} p_{i} \log \frac{p_{i}}{q_{i}}
    \label{eqn:ckld}
\end{equation}
where $p_{i}$ is the ratio of ground truth choice label for option ID $i$ and $q_{i}$  is the ratio for predictions.


\subsection{Investigation into the cause of Selection Bias in LLMs}
In the decoder-only transformer architecture, which is prevalent in most large language models (LLMs), each token is generated based on causal attention. This causal attention mechanism ensures that predictions are conditioned only on previously generated tokens. To preserve the sequential structure of the input, positional encodings are applied during attention computation.
When a question and its permuted-option variant are provided as input, the set of unique tokens remains unchanged. However, the reordering alters the positional encodings assigned to each option token. Since positional encodings influence attention scores, this modification can lead to differences in the model’s output, even if the semantic token content remains the same (see Appendix \ref{appendix:sensitivity} and Appendix \ref{appendix:investigation}).  




\subsection{Bias Mitigation Methodology}

In the following sections, we explain our efficient BaQCKV approach and the unsupervised LoRA-1 debiasing.  BaQCKV, an efficient majority voting variant that enforces permutation invariance (zero bias) with reduced compute via batched inference, ideal for critical evaluations requiring strict consistency; and LoRA-1, a lightweight adaptation method trained using our unsupervised bias metric to debias models for single-pass inference, suitable for large-scale deployments. BaQCKV trades computational overhead for robustness, while LoRA-1 prioritises scalability with minimal latency. This can enable  users to balance bias mitigation against operational constraints.  
\subsubsection{Efficient Majority Voting with BaQCKV}
The majority voting is an effective mitigation strategy for selection bias \cite{zong2023fool}.  It passes all option permutations of a question through the model and chooses the option with the highest average score across the permutations. This scheme enforces permutation invariance (0 bias on our metric) by  ensuring that an option has the same confidence across all permutations, making it an ideal selection bias mitigation strategy. Mathematically, majority voting calculates
\begin{equation}
i^* = \operatorname*{argmax}_{i \in O} \mathbb{E}_{\pi}\left[P(o_{\pi(i)} \mid Q, O_{\pi})\right],
\label{eqn:majority_voting}
\end{equation}
where \( O = \{o_1, \dots, o_n\} \) is the set of answer options, \( \pi \) denotes a permutation of the options, and \( \mathbb{E}_{\pi} \) represents the average over all \( n! \) permutations (as defined in Section~\ref{sec:select_bias_qant}). 

In spite of its effectiveness in mitigating bias, the majority voting has not been widely adopted, as the computational complexity of making predictions on all possible permutations is \(n!\) for an MCQ with \(n\) options. This can be easily reduced by defining a fixed number \(k\) and considering only \(k\) permutations instead of \(n!\), reducing the computational cost \citep{guda2024qmos}. Thus, \(p_i = \frac{1}{k}\sum_{j=1}^{k}p_{ji}\). However, this scheme can be made even more efficient, without a corresponding loss in bias, by employing a KV cache. To do so, we leverage the insight that while an MCQ consists of a question  \(Q\) (with or without a context), and a set of options, \(O\), \textit{the question, \(Q\), remains the same across all possible option permutations.} 
\newline
For $k$ permutations of the options, the original formulation of majority voting \citep{guda2024qmos} requires $k$ passes through the LLM, resulting in an additional overhead of $(k - 1) \times |\text{Questions} \oplus \text{Context} \oplus \text{Options}|$ tokens per question.
 We however, note that the set of $Questions \oplus Context$ tokens remains constant across all $k$ passes for each question in a batch. To eliminate the redundant computation of these tokens across the batch, we are motivated by the KV cache in \citep{pope2023efficiently}  to introduce the BaQCKV, which caches and reuses the KV states of the $\text{Questions} \oplus \text{Context}$ tokens for a set of $k$ permutations. This cached KV state is prepended to the KV states of the $k$ permuted options. The attention mask of the permuted options is then expanded based on the length of the $Questions \oplus Context$ tokens to ensure that the LLM's attention is correctly computed.
\begin{algorithm}[ht!]
\caption{Efficient Majority Inference with BaQCKV}
\label{alg:baqck}
\scriptsize
\begin{algorithmic}[1]
\Procedure{BaQCKVInference}{$Q_C, O_k, \mathcal{M}$}
    \State \textbf{Input:} $Q_C$ - Question $\oplus$ Context tokens, $O_k$ - $k$ permutations of options,  $\mathcal{M}$ - Language Model,  \textbf{Output:} $\mathcal{Y}_k$ - Model outputs
    \State
    
    \State \textbf{Step 1: Cache Question-Context KV States}
    \State $\text{\texttt{KV}}_{Q_C} \gets \mathcal{M}.\text{encode}(Q_C)$

    \State \textbf{Step 2: Compute KV States for Permuted Options}
    \For{$i = 1$ to $k$} \State $\text{\texttt{KV}}_{O_i}, \text{\texttt{mask}}_i \gets \mathcal{M}.\text{encode}(O_i)$ \EndFor

    \State \textbf{Step 3: Merge and Adjust KV States}
    \For{$i = 1$ to $k$} \State $\text{\texttt{KV}}_i \gets \text{\texttt{KV}}_{Q_C} \oplus \text{\texttt{KV}}_{O_i}, \quad \text{\texttt{mask}}_i \gets \mathbf{1}_{|Q_C|} \oplus \text{\texttt{mask}}_i$ \EndFor

    \State \textbf{Step 4: Compute Batch Outputs}
\State $\mathcal{Y}_k \gets \{\mathcal{M}.\text{decode}(\text{\texttt{KV}}_i, {\texttt{mask}}_i ) \mid i = 1, 2, \dots, k\}$
    \State \Return $\mathcal{Y}_k$
\EndProcedure
\end{algorithmic}
\end{algorithm}
We show in Appendix \ref{appendix:tokensavings} that the percentage of tokens saved by using the BaQCKV is defined by Equation (\ref{eqn:tknsave}), where  $C$ is the  optional set of context tokens for the Question $Q$.
\begin{equation}\label{eqn:tknsave}
\text{Token savings (\%)} = \frac{(k - 1) \times |Q \oplus C|}{k \times |Q \oplus C \oplus O|} \times 100 
\end{equation}
In Equation (\ref{eqn:tknsave}), the savings are maximized when $|C|$ is large, as in Retrieval-Augmented Generation (RAG), where redundant computation is minimized. Even when $|C| = 0$, savings persist due to the shared $|Q|$ tokens. Larger permutation sizes $k$ further amplify savings because of the increased redundancy in $|Q \oplus C|$ across permutations. Thus, BaQCKV is most effective in tasks with substantial shared context, multiple options, and large permutation sizes. 

\subsubsection{Unsupervised LoRA-1 Bias Mitigation}\label{sec:LoRA1-1}
We introduce an unsupervised fine-tuning of \textbf{PBM}, our permutation-based bias metric, to mitigate the selection bias. This is because PBM is fully differentiable unlike the Fluctuation Rate  (FR) and the Standard Deviation of Recalls (RStd). In addition, it is also label-free, unlike all the other metrics including the   CKLD. We make two adjustments to the metric when using it as a loss for fine-tuning (equation (\ref{eqn:loss})). Firstly, to ensure that there is an adequate flow of information from the gradients we take the variance of the log of the probabilities (Equation \eqref{eqn:bias_log}). When obtaining the option probabilities from the model, we only consider the logits that correspond to the option IDs instead of logits for the entire vocabulary. Secondly, we observe that a model can learn to minimize the bias by simply predicting a uniform probability for all options IDs across all permutations. In that case, the mean probability for all option IDs would be the same as the uniform probabilities assigned to all option IDs for all permutations. To prevent this, we regularize the bias with the entropy across the option IDs (Equation \eqref{eqn:entropy}). This helps in making the model maintain its confidence in the chosen answer while also minimizing the bias across different permutations. $\lambda$ is a hyper-parameter that balances  the model's confidence in an answer and minimizing the bias. Computing the loss defined in Equation \eqref{eqn:loss} can be computationally expensive, since each example must be expanded to all possible permutations. To reduce this cost, first, we apply the BaQCKV to compute the loss and only sample a maximum of 24 permutations for questions with more than 4 options.
\begin{equation}
\text{Loss}= B(Q, O)_{log} + \lambda H(Q,O)
 \label{eqn:loss}
\end{equation}
\newline
where
\begin{equation}
\begin{split}
    B(Q, O)_{\text{log}} &= \sum_{\pi} \biggl( \log \left( P(o_{\pi(i)} \mid Q, O_{\pi}) \right) \\
    &\quad - \log \left( \mathbb{E}_{\pi} \left[ P(o_{\pi(i)} \mid Q, O_{\pi}) \right] \right) \biggr)^2
\end{split}
\label{eqn:bias_log}
\end{equation}

\begin{equation}
 H(Q,O) =   -\sum_{\pi} \sum_{i} (P(o_{\pi(i)} \log (P(o_{\pi(i)}))
 \label{eqn:entropy}
 \end{equation}
As LLMs are desired to be used for a wide variety of tasks and not just answering MCQs, we adopt the LoRA fine-tuning \cite{hu2021loralowrankadaptationlarge} to preserve the original performance of the LLM on non-MCQ tasks while avoiding expensive training. The LoRA debiasing weight adapters can be connected when the model is used for MCQ. 

\subsubsection{Complexity Analysis}
We present a comparison of the computational requirements of various bias metrics in  Table \ref{tab:complexity_metrics}, focusing on the number of permutations each metric considers and the effective token cost passed through the model. Existing metrics such as RStd, CKLD, and FR operate over one or two fixed permutations and require ground-truth labels, limiting both their expressiveness and flexibility. In contrast, our proposed PBM evaluates how model confidence varies across different permutations of answer options, offering a more comprehensive and label-free assessment of positional bias.
While full PBM evaluation over all $n!$ permutations is computationally expensive, we use both two permutation sampling and BaQCKV to make PBM scalable. Sampling allows us to approximate the metric using only $m \ll n$ permutations, and BaQCKV further reduces cost by reusing the shared context $ Q \oplus C $, resulting in a total complexity of $ \mathcal{O}(|Q \oplus C| + m|O|) $ tokens per query.
This approach strikes an effective balance between computational efficiency and expressive power.
\begin{table}[ht]
\centering
\small
\setlength{\tabcolsep}{2pt} 
\renewcommand{\arraystretch}{1.1} 
\begin{tabular}{@{}lccc@{}}
\toprule
\textbf{Metric} & \textbf{Labels?} & \textbf{Perms} & \(\mathcal{O}(\cdot)\) \\
\midrule
RStd        & Yes & 1               & \(|Q \oplus C \oplus O | \) \\
CKLD        & Yes & 1               &  \(|Q \oplus C \oplus O| \)\\
FR          & No  & 2               & \(2 |Q \oplus C \oplus O| \) \\
PBM (full)  & No  & \(n!\)          & \(n!|Q \oplus C \oplus O|\) \\
\quad + sampling & No & \(m \ll n!\) & \(m(|Q \oplus C \oplus O|)\) \\
\quad + BaQCKV & No & \(m \ll n!\) & \( |Q \oplus C|  + m|O|\) \\
\bottomrule
\end{tabular}
\caption{Computational comparison of bias metrics.}
\label{tab:complexity_metrics}
\end{table}

\section{Results}\label{sec:results}
\textbf{Datasets and Models :} For our experiments, we employed three small language models of comparable size: Qwen2.5-3B-Instruct \citep{bai2023qwentechnicalreport}, Phi-2 \citep{javaheripi2023phi}, and Llama3.2-3B \citep{grattafiori2024llama3herdmodels}. We experiment with these models on four diverse datasets across different domains: TeleQnA \citep{maatouk2023teleqnabenchmarkdatasetassess}, MedMCQA \citep{pal2022medmcqalargescalemultisubject}, QASC\citep{khot2020qascdatasetquestionanswering} and ARC Challenge\citep{Clark2018ThinkYH}. ARC Challenge and MedMCQA have 4 options and TeleQnA has 2-5 options, while  QASC has 8. A full description of these datasets and their statistics can be found in Appendix~\ref{appendix:dataset-description}

\textbf{Experiments:}
We conduct experiments to evaluate both the accuracy and bias of various models across all datasets, and to assess the effectiveness of different selection bias mitigation strategies. First, we demonstrate that \textit{majority voting not only reduces or eliminates selection bias (quantified by PBM) but also improves model accuracy}. Furthermore, by introducing BaQCKV, we show that \textit{majority voting can be made significantly more efficient}, yielding substantial savings in both computation time and token usage. In addition, we assess the impact of our proposed \textbf{LoRA-1 fine-tuning method}, which, on average, reduces the variation of accuracy, PBM and Fluctuation Rate and increases accuracy across all datasets/models and exhibits strong transferability across datasets.

We compare our approach against three alternative bias mitigation methods: (1) \mbox{\textbf{GRAY} \cite{wei-etal-2024-unveiling}}, a gray-box technique that leverages both forward and backward predictions to reduce bias; (2) \mbox{\textbf{BNP (Bias Node Pruning)}} \cite{choi2024mitigatingselectionbiasnode}, which prunes parameters in the final projection layer that contribute to bias; and (3) \mbox{\textbf{PRIDE} \cite{zheng2024large}}, which normalizes model predictions using prior probabilities estimated from the dataset.

\textbf{However, these prior methods do not achieve the consistency of majority voting, a technique well-established for mitigating selection bias} \cite{zong2023fool,selfconsistency2024}. The prompt templates and hyperparameter configurations used in our experiments are documented in Appendix ~\ref{appendix:prompt-template}.

\begin{table*}[h!]
\centering
\resizebox{\textwidth}{!}{%
\begin{tabular}{lcccccccccccccccc}
\multicolumn{1}{c}{\bf Model Name} & \multicolumn{4}{c}{\bf TeleQnA} & \multicolumn{4}{c}{\bf MedMCQA} & \multicolumn{4}{c}{\bf QASC} & \multicolumn{4}{c}{\bf ARC} \\
\cmidrule(lr){2-5} \cmidrule(lr){6-9} \cmidrule(lr){10-13} \cmidrule(lr){14-17} 
& \textbf{Acc} & \textbf{PBM} & \textbf{TimeS (\%)} & \textbf{TokS (\%)} 
& \textbf{Acc} & \textbf{PBM} & \textbf{TimeS (\%)} & \textbf{TokS (\%)} 
& \textbf{Acc} & \textbf{PBM} & \textbf{TimeS (\%)} & \textbf{TokS (\%)} 
& \textbf{Acc} & \textbf{PBM} & \textbf{TimeS (\%)} & \textbf{TokS (\%)} \\
\hline
Qwen2.5-3B & 0.5464 & 0.021 & - & - & 0.479 & 0.058 & - & - & 0.737 & 0.011 & - & - & 0.804 & 0.029 & - & - \\
Qwen2.5-3B + MV & 0.5710 & 0.000 & 0.9513 & 0.4583 & 0.487 & 0.000 &  0.9306 & 0.5328 & 0.947 & 0.000 & 0.8432 & 0.5377 & 0.839 & 0.00 & 0.9292 & 0.3600 \\
\midrule
Phi-2 & 0.2568 & 0.0303 & - & - & 0.359 & 0.082 & - & - & 0.630 & 0.024 & - & - & 0.552 & 0.0269 & - & - \\
Phi-2 + MV & 0.328 & 0.000 & 0.8919 & 0.3596 & 0.369 & 0.000 &0.9351 & 0.551 & 0.9329 & 0.000 & 0.9333 & 0.543 & 0.4547 & 0.000 & 0.9316 & 0.366 \\
\midrule
Llama3.2-3B & 0.4536 & 0.0053 & - & - & 0.370 & 0.017 & - & - & 0.4892 & 0.005 & - & - & 0.5179 & 0.0091 & - & - \\
Llama3.2-3B + MV & 0.516 & 0.000 & 0.9340 & 0.4618 & 0.384 & 0.000 & 0.9352 & 0.533 & 0.837 & 0.000 & 0.9482 & 0.545 & 0.537 & 0.00 & 0.9142 & 0.3639 \\
\end{tabular}%
}
\caption{Accuracy and bias values for different models across multiple datasets, along with computational efficiency improvements using Majority Voting (MV) with BaQCKV. TimeS is the percentage of inference time speedup as measured, and TokS is the token saved, respectively.}
\label{tab:results}
\end{table*}
\textbf{Our unsupervised bias metric (PBM) correlates with the difficulty of the MCQ:} The results in Table \ref{tab:results} show that all models exhibit varying degrees of bias, correlating with the difficulty of the problem. Across all models, the bias is seen to be highest with the MedMCQA benchmark due to its difficulty (having the lowest accuracy). This confirms that selection bias is present and measurable using our proposed metric (PBM). This also means that PBM may be used to compare the difficulty of different MCQ  datasets without having access to the labels.   Notably, after applying majority voting (MV) with the help of BaQCKV, the PBM value consistently drops to 0.00.  Additionally, applying majority voting shows substantial improvements in accuracy, particularly in QASC, where scores increase significantly (e.g., from 0.630 to 0.9329 for Phi-2 and 0.4892 to 0.837 for Llama), validating the effectiveness of our metric in capturing and mitigating bias.

\textbf{Efficiency of the BaQCKV:} Beyond bias reduction, the BaQCKV enhances real-world applicability by significantly reducing the computational costs of applying majority voting. As shown in Table \ref{tab:results}, our efficient BaQCKV approach for the majority voting results in significant time savings of up to 84\% (Qwen on QASC) and 89\%  ( Phi on TeleQnA) and over 90\% across token savings across all models and datasets. This efficiency gain is crucial for deploying bias-mitigation strategies at scale during inference, making our approach feasible for real-world applications where computational cost is a limiting factor. 

\textbf{The unsupervised LoRA-1 Bias Mitigation demonstrates the best performance in maintaining consistency:}
\begin{table}[t]
\centering
\scriptsize
\setlength{\tabcolsep}{4pt}
\begin{tabular}{lrrrrrr}
\toprule
Method & \textbf{PBM} $\downarrow$ & RStd $\downarrow$ & CKLD $\downarrow$ & FR $\downarrow$ & Acc $\uparrow$  & AccStd $\downarrow$ \\
\midrule
LoRA-1   & -0.586 & -0.076 & 0.928 & -0.525 & 0.200 & \textbf{-0.276} \\
Gray   & -0.364 & -0.045 & 0.677 & --     & 0.077 & 0.940 \\
BNP     & -0.119 &  0.000 & 0.653 & -0.250 & 0.064 & -0.131 \\
PriDe  &  1.880 & -0.240 & \textbf{0.432} & -0.137 & \textbf{0.419} & -0.040\\
\bottomrule
\end{tabular}
\vspace{-0.1in}
\caption{Comparison of mitigation methods on all datasets and models. The values are the average percentage change of biases, accuracy and standard deviation (std) of accuracy of the debiased model over the original models. The Acc here is the change in accuracy using the original permutation while the AccStd measures the std of accuracy across different permutations.} 
\label{tab:mitigation_results}
\end{table}

\begin{figure}[t]
    \centering
    \begin{subfigure}[t]{0.95\linewidth}
        \centering
        \includegraphics[width=\linewidth]{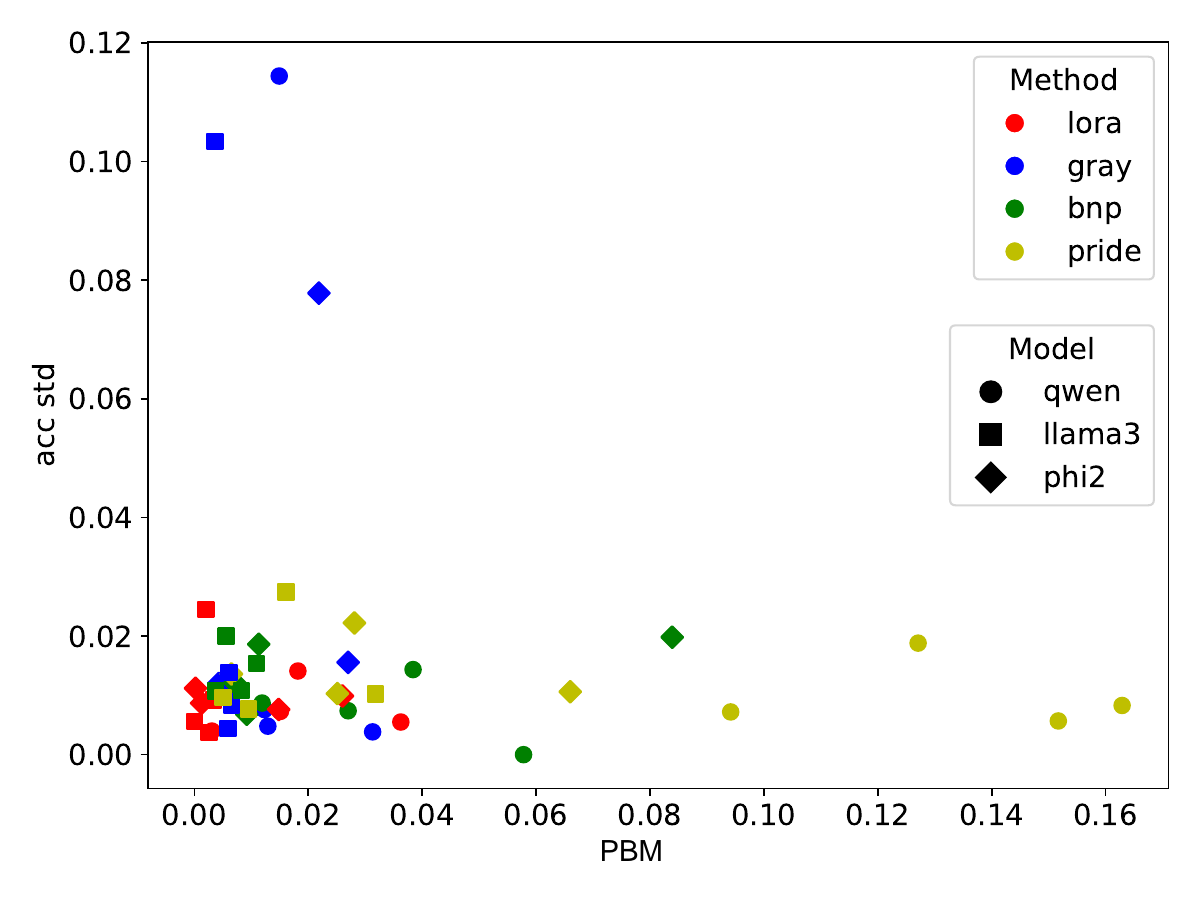}
        \caption{Standard deviation of accuracy across permutations vs bias across all models and datasets for different mitigation strategies.}
        \label{fig:acc_std-bias}
    \end{subfigure}
    \begin{subfigure}[t]{\linewidth}
        \centering
        \includegraphics[width=\linewidth]{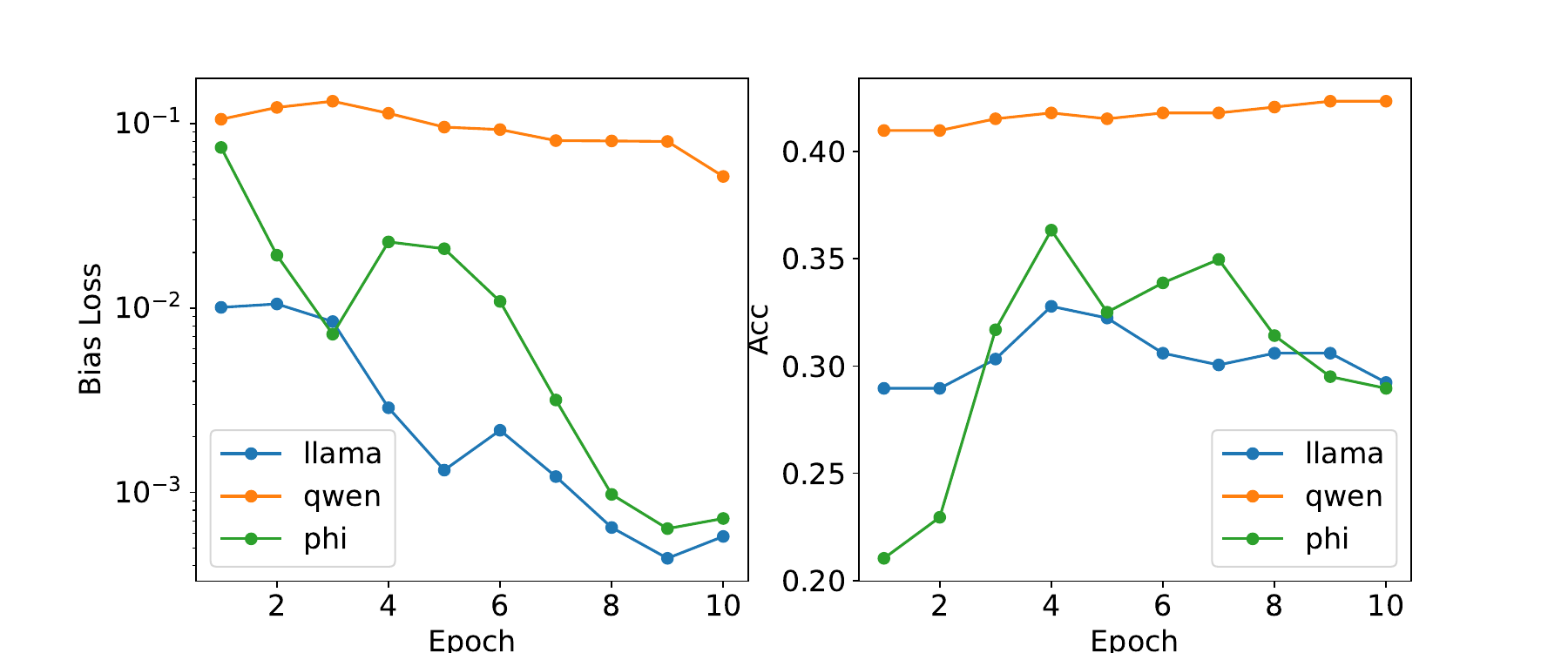}
        \caption{LoRA1 finetuning training graph showing bias and accuracy for the TeleQnA dataset.}
        \label{fig:finetuning-teleqna}
    \end{subfigure}
    \caption{Visualization of bias-related behaviors across models and strategies.}
    \label{fig:bias-figs}
\end{figure}

The scatter plot in Figure~\ref{fig:acc_std-bias} illustrates that, overall, the \texttt{LoRA-1} fine-tuning approach exhibits greater consistency in accuracy, characterised by a smaller standard deviation and lower selection bias \textbf{(PBM)} compared to other mitigation strategies. Ideally, an effective model should have its corresponding points converge near the origin of the plot, indicating minimal variability and bias. Moreover, as shown in Table ~\ref{tab:mitigation_results}, when evaluating the percentage change relative to the undebiased models, \texttt{LoRA-1} fine-tuning achieves the greatest average reduction in both standard deviation of accuracy ($-27\%$), PBM bias ($-58\%$) and FR ($-52\%$) across all evaluated models and datasets. These results highlight the effectiveness of \texttt{LoRA-1} in mitigating variability and bias simultaneously. It also, on average, improves the accuracy of the models by 20\%  even though this is not up to the 41.92\% demonstrated by PriDe. Also, PriDe shows the highest reduction in the RStd ($-24\%$) and the smallest increase in CKLD ($43.2\%$). However, PriDe is not helpful in reducing the PBM bias, it rather increases  PBM bias ($180\%$). It also hardly offers any improvements in the standard deviation of accuracy ($-0.4\%$). We show a training graph of the unsupervised finetuning for the TeleQnA dataset process in Figure \ref{fig:finetuning-teleqna}.



\begin{table}[t]
\centering
\scriptsize
\setlength{\tabcolsep}{3pt}
\begin{tabular}{lrrrrrr}
\toprule
Model-Dataset-Train & PBM $\downarrow$ & RStd $\downarrow$ & CKLD  $\downarrow$ & FR $\downarrow$ & Acc $\uparrow$& AccStd $\downarrow$ \\
\midrule
QWEN-MEDCQ      & -0.497 & -0.411 & 207.74 & -0.150 & 0.028 & -0.279 \\
QWEN-TeleQNA    & -0.509 &  0.710 &   4.01 & -0.160 & 0.029 & -0.206 \\
QWEN-ARC        & -0.413 &  1.017 &   4.63 & -0.126 & 0.026 & -0.234 \\
QWEN-QASC       & -0.319 &  1.168 &   5.06 & -0.006 & -0.000 & -0.153 \\
Phi2-ARC        & -0.640 & -0.712 &  -0.90 & -0.764 & 0.423 & -0.371 \\
Llama3.2-TeleQNA & -0.687 & -0.425 &  -0.60 & -0.760 & 0.282 & -0.479 \\
Llama3.2-ARC    & -0.799 & -0.314 &   0.18 & -0.651 & 0.220 & -0.300 \\
Llama3.2-QASC   & -0.863 &  0.213 &   0.54 & -0.627 & -0.085 & -0.073 \\
\bottomrule
\end{tabular}
\vspace{-0.1in}
\caption{The average percentage change in bias metrics and accuracy for transferability experiments. For each row, the model-dataset is evaluated on all other datasets excluding the one used for finetuning}
\label{tab:transfer_results}
\end{table}


\textbf{The transferability of the unsupervised  LoRA-1 Bias Mitigation:}
We investigate the transferability of the unsupervised LoRA-1 fine-tuning approach by training a model on a single dataset and evaluating the resulting checkpoint on all other datasets. The average performance across each model/dataset pair is reported in Table~\ref{tab:transfer_results}, with complete results provided in Appendix~\ref{app:transfer}. As shown in Table~\ref{tab:transfer_results}, the unsupervised fine-tuning generally transfers well: it consistently reduces our permutation bias metric (PBM), fluctuation rate, and the standard deviation of accuracy across option permutations. In many cases, it also yields modest improvements in accuracy. However, similar to the non-transfer setting, this approach does not improve the CKLD metric. Notably, for the Qwen-MEDCQ model, CKLD actually increases significantly, accompanied by a more than 200\% rise in RStd. This may be due to the model shifting toward greater consistency across permutations, which can lead to more uniform confidence distributions that diverge from dataset-specific label frequencies. Importantly, this behavior aligns with our objective of reducing positional sensitivity. However, future studies will need to be done to properly analyse and understand this behaviour.

We compare the computational efficiency of the different bias mitigation strategies in Appendix ~\ref{appendix:computational-efficiency}.

\section{Conclusion}\label{sec:conclusion}
In this work, we address a critical yet underexplored challenge in Multiple-Choice  (MCQ)  Question Answering—selection bias in Large Language Models (LLMs). We introduced a novel unsupervised, label-free bias metric (PBM) that directly quantifies inconsistencies in predictions across permuted answer choices, offering a more faithful measure of selection bias than existing methods. To mitigate this bias without incurring prohibitive computational costs, we proposed BaQCKV, an efficient majority voting strategy, and LoRA-1, a lightweight fine-tuning method grounded in our bias metric. Our experiments across diverse MCQ datasets and models demonstrate that these techniques not only reduce bias and improve accuracy but also significantly cut down inference time and token usage, making them scalable and practical for real-world deployment. Ultimately, our work provides both a theoretical and practical framework for more reliable and efficient MCQ with LLMs.
\section{Limitations}\label{sec:limitations}
This work only focuses on decoder only transformer language models and did not investigate bias in other language models such as encoder-decoder models. Also, we only investigate MCQs where the model has to choose one option and do not consider other types of MCQs.

In practice, multiple-choice questions typically have only 2–5 options. Thus, we fix the maximum at 24 permutations, which we believe balances computational costs with practical applicability. MCQs with more options are rarer. This ensures that the metric remains both meaningful and efficient in real-world MCQ settings. Also, we show that using these 24 permutations is sufficient, because it does well on the QASC dataset that has 8 options and the TeleQnA that has 2-5 options. However, future works may consider more permutations and larger LLM models.

\bibliography{bibfile}

\onecolumn
\appendix
\section{Appendix}

\subsection{Positional Encoding and Sensitivity of  Transformers to Option Permutations}\label{appendix:sensitivity}

In decoder-only transformers, token generation is conditioned on causal self-attention, where each token attends to prior tokens using both content-based embeddings and positional encodings. When a question \( Q \) and its associated options \( O = \{o_1, o_2, \dots, o_n\} \) are presented, the model processes the sequence:
\[
S = Q \oplus O.
\]
For a permutation \( \pi \) of the options, the modified sequence becomes:
\[
S_{\pi} = Q \oplus O_{\pi}.
\]

While the token set remains unchanged, the reordering affects positional encodings, altering attention computations. The self-attention mechanism computes attention scores between tokens at positions \( i \) and \( j \) as:
\[
\text{Attention}_{i,j} = \frac{Q (o_i + p_i) \cdot (K(o_j + p_j))^T}{\sqrt{d}}.
\]
For the permuted sequence \( S_{\pi} \), the updated scores are:
\[
\text{Attention}_{i,j}^{\pi} = \frac{Q (o_i + p_{\pi(i)}) \cdot (K(o_j + p_{\pi(j)}))^T}{\sqrt{d}}.
\]
Since \( p_i \neq p_{\pi(i)} \), the attention patterns for \( S \) and \( S_{\pi} \) differ, resulting in distinct contextual representations for the same token set.

As attention weights directly influence token representations, these changes propagate through the network, modifying the sequence representation and ultimately affecting the model’s output distribution. Let \( P(y \mid S) \) and \( P(y \mid S_{\pi}) \) denote the probability distributions over possible answers. Then, 
\[
P(y \mid S) \neq P(y \mid S_{\pi}).
\]

\subsection{Impact of option permutations on attention scores}\label{appendix:investigation}
\begin{figure}[H]
\includegraphics[width=0.9\linewidth]{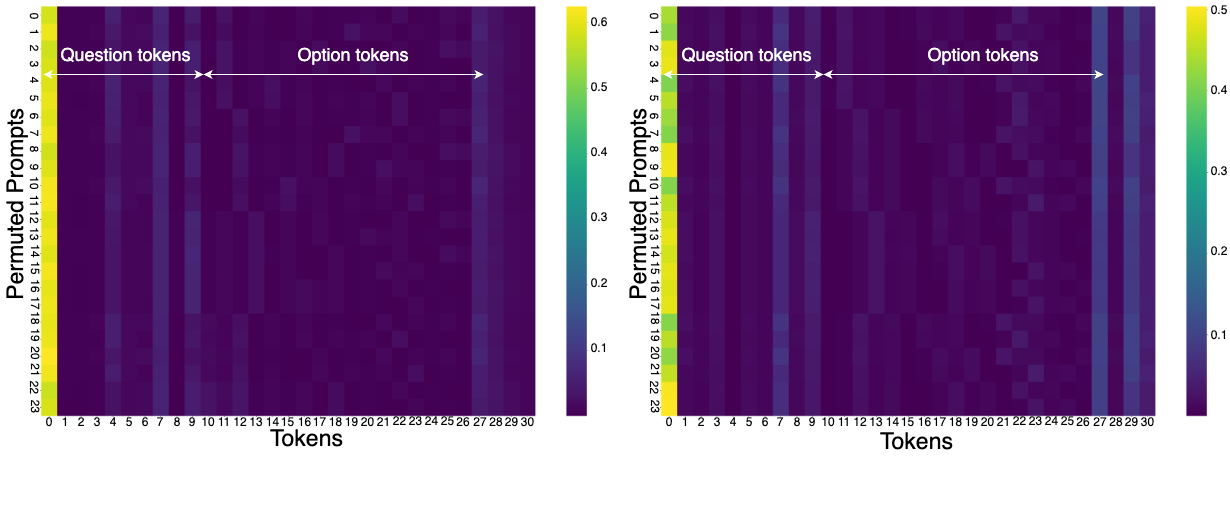}
  \caption {Attention scores for the last token in the last layer of the Llama model across different prompt permutations, shown for two transformer heads.}
    \label{fig:cause_bias}
\end{figure}

The impact of the permutations owing to the positional encoding on the attention scores is illustrated in Figure \ref{fig:cause_bias}. The figure shows that there are more fluctuations in the attention scores on the option tokens (later parts of the x-axis)  compared to the question tokens (earlier tokens on the x-axis). 

\subsection{Proof of Token Savings in BaQCKV} \label{appendix:tokensavings}

In the original Majority Voting (MV) framework, each question undergoes $k$ passes through the LLM, processing the full sequence of $Q \oplus C \oplus O$ each time. The total token cost per question is:
\begin{equation}
    \text{Cost}_{\text{MV}} = k \times |Q \oplus C \oplus O|
\end{equation}

In BaQCK, the shared $Q \oplus C$ tokens are processed only once, while the $O$ tokens are processed $k$ times. Thus, the total token cost per question is:
\begin{equation}
    \text{Cost}_{\text{MV}} = |Q \oplus C| + k \times |O|
\end{equation}

The token savings is computed as:
\begin{align}
    \text{Savings} &= \text{Cost}_{\text{MV}} - \text{Cost}_{\text{BaQCK}} \\
    &= k \times |Q \oplus C \oplus O| - (|Q \oplus C| + k \times |O|) \\
    &= k \times |Q \oplus C| + k \times |O| - |Q \oplus C| - k \times |O| \\
    &= (k - 1) \times |Q \oplus C|
\end{align}

Expressing this as a percentage of the original cost:
\begin{equation}
    \text{Token savings (\%)} = \frac{(k - 1) \times |Q \oplus C|}{k \times |Q \oplus C \oplus O|} \times 100
\end{equation}

This result shows that BaQCK significantly reduces token computations, particularly when $|C|$ is large (e.g., in Retrieval-Augmented Generation). Even for small or zero-context cases ($|C| = 0$), savings persist due to shared $|Q|$ tokens. Increasing $k$ further amplifies efficiency by reducing redundant recomputation across shuffled options.

\section{Experimental Results}
\subsection{Bias Mitigation Results on all models and datasets}
\begin{table}[h!]
\centering
\small
\caption{QWEN Performance across Datasets (Raw Scores)}
\begin{tabular}{@{}lcccccc@{}}
\toprule
\textbf{Metric} & \textbf{PBM} & \textbf{RStd} & \textbf{CKLD} & \textbf{Fluct. Rate} & \textbf{Accuracy} & \textbf{Perm Acc Std} \\
\midrule
\multicolumn{7}{c}{\textit{ARC Dataset}} \\
\midrule
Baseline & 0.0268 & 0.0158 & 0.0003 & 0.1997 & 0.8038 & 0.0088 \\
LoRA FineTuning & 0.0151 & 0.0101 & 0.0004 & 0.1869 & 0.808 & 0.0073 \\
Gray-Box Weighting & 0.0129 & 0.0083 & 0.0002 & -- & 0.8283 & 0.0048 \\
BNP & 0.0270 & 0.0169 & 0.0004 & 0.2550 & 0.7994 & 0.0074 \\
Pride & 0.1629 & 0.0261 & 0.0007 & 0.2374 & 0.8039 & 0.0083 \\
\midrule
\multicolumn{7}{c}{\textit{TeleQNA Dataset}} \\
\midrule
Baseline & 0.0214 & 0.0564 & 0.0070 & 0.5874 & 0.5464 & 0.0189 \\
LoRA FineTuning & 0.0182 & 0.0879 & 0.0223 & 0.6093 & 0.5437 & 0.0141 \\
Gray-Box Weighting & 0.0149 & 0.0179 & 0.0070 & -- & 0.5492 & 0.1144 \\
BNP & 0.0384 & 0.0881 & 0.0145 & 0.2158 & 0.5546 & 0.0143 \\
Pride & 0.1271 & 0.0778 & 0.0040 & 0.3389 & 0.4186 & 0.0188 \\
\midrule
\multicolumn{7}{c}{\textit{MedMCQ Dataset}} \\
\midrule
Baseline & 0.0577 & 0.0179 & 0.0006 & 0.8964 & 0.4805 & 0.0057 \\
LoRA FineTuning & 0.0363 & 0.0771 & 0.0082 & 0.9031 & 0.4798 & 0.0055 \\
Gray-Box Weighting & 0.0313 & 0.0087 & 0.0062 & 0.0000 & 0.4867 & 0.0038 \\
BNP & 0.0578 & 0.0096 & 0.0056 & 0.2421 & 0.4797 & 0.0000 \\
Pride & 0.1517 & 0.0238 & 0.0078 & 0.2476 & 0.4801 & 0.0057 \\
\midrule
\multicolumn{7}{c}{\textit{QASC Dataset}} \\
\midrule
Baseline & 0.0110 & 0.0547 & 0.0117 & 0.2062 & 0.8952 & 0.0068 \\
LoRA FineTuning & 0.0031 & 0.0101 & 0.0003 & 0.1188 & 0.9698 & 0.0040 \\
Gray-Box Weighting & 0.0123 & 0.0324 & 0.0146 & -- & 0.8801 & 0.0076 \\
BNP & 0.0119 & 0.0573 & 0.0129 & 0.1619 & 0.8866 & 0.0087 \\
Pride & 0.0942 & 0.0594 & 0.0060 & 0.1690 & 0.8729 & 0.0072 \\
\bottomrule
\end{tabular}
\end{table}

\begin{table}[H]
\centering
\small
\caption{Phi-2 Performance across Datasets (Raw Scores)}
\begin{tabular}{@{}lcccccc@{}}
\toprule
\textbf{Metric} & \textbf{PBM} & \textbf{RStd} & \textbf{CKLD} & \textbf{Fluct. Rate} & \textbf{Accuracy} & \textbf{Perm Acc Std} \\
\midrule
\multicolumn{7}{c}{\textit{ARC Dataset}} \\
\midrule
Baseline & 0.0269 & 0.2207 & 0.2853 & 0.6334 & 0.5520 & 0.0110 \\
LoRA FineTuning & 0.0214 & 0.2102 & 0.2501 & 0.6225 & 0.5589 & 0.0082 \\
Gray-Box Weighting & 0.0187 & 0.1965 & 0.2394 & -- & 0.5661 & 0.0071 \\
BNP & 0.0276 & 0.2251 & 0.2923 & 0.6881 & 0.5498 & 0.0104 \\
Pride & 0.1507 & 0.2769 & 0.2933 & 0.6987 & 0.5512 & 0.0107 \\
\midrule
\multicolumn{7}{c}{\textit{TeleQNA Dataset}} \\
\midrule
Baseline & 0.0303 & 0.3559 & 1.5208 & 0.8114 & 0.2568 & 0.0216 \\
LoRA FineTuning & 0.0255 & 0.3471 & 1.4801 & 0.8041 & 0.2642 & 0.0173 \\
Gray-Box Weighting & 0.0196 & 0.3432 & 1.4273 & -- & 0.2697 & 0.0152 \\
BNP & 0.0388 & 0.3674 & 1.5404 & 0.7463 & 0.2553 & 0.0201 \\
Pride & 0.1301 & 0.3865 & 1.4311 & 0.7522 & 0.2374 & 0.0214 \\
\midrule
\multicolumn{7}{c}{\textit{MedMCQ Dataset}} \\
\midrule
Baseline & 0.0512 & 0.3981 & 1.4585 & 0.9369 & 0.3409 & 0.0194 \\
LoRA FineTuning & 0.0418 & 0.3907 & 1.4132 & 0.9286 & 0.3467 & 0.0168 \\
Gray-Box Weighting & 0.0356 & 0.3849 & 1.4017 & 0.0000 & 0.3523 & 0.0150 \\
BNP & 0.0542 & 0.4072 & 1.4682 & 0.8624 & 0.3397 & 0.0175 \\
Pride & 0.1489 & 0.4138 & 1.4425 & 0.8796 & 0.3402 & 0.0189 \\
\midrule
\multicolumn{7}{c}{\textit{QASC Dataset}} \\
\midrule
Baseline & 0.0314 & 0.3232 & 2.9914 & 0.9676 & 0.1577 & 0.0122 \\
LoRA FineTuning & 0.0249 & 0.3014 & 2.7412 & 0.9548 & 0.1643 & 0.0103 \\
Gray-Box Weighting & 0.0221 & 0.2876 & 2.6785 & -- & 0.1671 & 0.0094 \\
BNP & 0.0327 & 0.3357 & 3.0141 & 0.9251 & 0.1552 & 0.0109 \\
Pride & 0.1273 & 0.3428 & 2.8345 & 0.9387 & 0.1499 & 0.0118 \\
\bottomrule
\end{tabular}
\end{table}

\begin{table}[H]
\centering
\small
\caption{LLama 3.2 Performance across Datasets (Raw Scores)}
\begin{tabular}{@{}lcccccc@{}}
\toprule
\textbf{Metric} & \textbf{PBM} & \textbf{RStd} & \textbf{CKLD} & \textbf{Fluct. Rate} & \textbf{Accuracy} & \textbf{Perm Acc Std} \\
\midrule
\multicolumn{7}{c}{\textit{ARC Dataset}} \\
\midrule
Baseline & 0.0091 & 0.1082 & 0.1050 & 0.5171 & 0.5179 & 0.0097 \\
LoRA FineTuning & 0.0034 & 0.1780 & 0.1144 & 0.1373 & 0.5339 & 0.0092 \\
Gray-Box Weighting & 0.0065 & 0.0239 & 0.0999 & 0.0000 & 0.5957 & 0.0083 \\
BNP & 0.0083 & 0.0500 & 0.0230 & 0.3279 & 0.5392 & 0.0108 \\
Pride & 0.0318 & 0.0911 & 0.0125 & 0.4789 & 0.5615 & 0.0102 \\
\midrule
\multicolumn{7}{c}{\textit{TeleQNA Dataset}} \\
\midrule
Baseline & 0.0053 & 0.1010 & 0.0907 & 0.5574 & 0.4536 & 0.0165 \\
LoRA FineTuning & 0.0002 & 0.1998 & 0.3120 & 0.3306 & 0.3830 & 0.0245 \\
Gray-Box Weighting & 0.0036 & 0.0330 & 0.0828 & --     & 0.4004 & 0.1034 \\
BNP & 0.0056 & 0.1786 & 0.0614 & 0.1746 & 0.4754 & 0.0199 \\
Pride & 0.0161 & 0.1638 & 0.0483 & 0.5575 & 0.4317 & 0.0274 \\
\midrule
\multicolumn{7}{c}{\textit{MedMCQ Dataset}} \\
\midrule
Baseline & 0.0167 & 0.3182 & 0.5597 & 0.8128 & 0.3696 & 0.0161 \\
LoRA FineTuning & 0.0000 & 0.0052 & 0.0531 & 0.2116 & 0.3213 & 0.0056 \\
Gray-Box Weighting & 0.0059 & 0.0221 & 0.8249 & --     & 0.4014 & 0.0044 \\
BNP & 0.0109 & 0.3098 & 0.5049 & 0.7547 & 0.3851 & 0.0154 \\
Pride & 0.0095 & 0.0165 & 0.0112 & 0.8217 & 0.3973 & 0.0077 \\
\midrule
\multicolumn{7}{c}{\textit{QASC Dataset}} \\
\midrule
Baseline & 0.0046 & 0.2021 & 0.4818 & 0.6544 & 0.4892 & 0.0138 \\
LoRA FineTuning & 0.0026 & 0.0130 & 0.0014 & 0.1263 & 0.9514 & 0.0037 \\
Gray-Box Weighting & 0.0061 & 0.2500 & 0.3122 & --     & 0.6210 & 0.0138 \\
BNP & 0.0037 & 0.1357 & 0.1357 & 0.3359 & 0.7441 & 0.0107 \\
Pride & 0.0050 & 0.0545 & 0.0017 & 0.4841 & 0.8715 & 0.0096 \\
\bottomrule
\end{tabular}
\end{table}

\subsection{Transferability Results over all datasets and models}\label{app:transfer}
\begin{table}[H]
\centering
\scriptsize
\caption{Transferability of Model-Checkpoints Across Datasets (Complete Results)}
\begin{tabular}{@{}lcccccc@{}}
\toprule
\textbf{Source/Target} & \textbf{PBM} & \textbf{RStd} & \textbf{CKLD} & \textbf{Fluct. Rate} & \textbf{Accuracy} & \textbf{Perm Acc Std} \\
\midrule
\multicolumn{7}{c}{\textbf{QWEN/MEDCQ $\rightarrow$ ARC Dataset}} \\
\midrule
Baseline & 0.0268 & 0.0158 & 0.0003 & 0.1997 & 0.8038 & 0.0088 \\
LoRA FineTune & 0.0101 & 0.0004 & 0.1869 & 0.1869 & 0.8080 & 0.0073 \\
\% Change & -43.6\% & -36.1\% & +333.3\% & -6.4\% & +0.5\% & -17.0\% \\
LoRA Transfer & 0.0215 & 0.0116 & 0.0001 & 0.2438 & 0.8034 & 0.0075 \\
\midrule
\multicolumn{7}{c}{\textbf{QWEN/MEDCQ $\rightarrow$ TeleQNA Dataset}} \\
\midrule
Baseline & 0.0214 & 0.0564 & 0.0070 & 0.5874 & 0.5464 & 0.0189 \\
LoRA FineTune & 0.0182 & 0.0879 & 0.0223 & 0.6093 & 0.5437 & 0.0141 \\
\% Change & -15.0\% & +55.9\% & +218.6\% & +3.7\% & -0.5\% & -25.4\% \\
LoRA Transfer & 0.0254 & 0.0671 & 0.0096 & 0.2076 & 0.5574 & 0.0116 \\
\midrule
\multicolumn{7}{c}{\textbf{QWEN/MEDCQ $\rightarrow$ QASC Dataset}} \\
\midrule
Baseline & 0.0110 & 0.0547 & 0.0117 & 0.2062 & 0.8952 & 0.0068 \\
LoRA FineTune & 0.0031 & 0.0101 & 0.0003 & 0.1188 & 0.9698 & 0.0040 \\
\% Change & -72.2\% & -81.5\% & -97.4\% & -42.4\% & +8.3\% & -41.2\% \\
LoRA Transfer & 0.0097 & 0.0720 & 0.0291 & 0.1847 & 0.8585 & 0.0071 \\
\bottomrule
\end{tabular}
\end{table}

\begin{table}[H]
\centering
\scriptsize
\caption{Transferability of Model-Checkpoints (Continued)}
\begin{tabular}{@{}lcccccc@{}}
\toprule
\textbf{Source/Target} & \textbf{PBM} & \textbf{RStd} & \textbf{CKLD} & \textbf{Fluct. Rate} & \textbf{Accuracy} & \textbf{Perm Acc Std} \\
\midrule
\multicolumn{7}{c}{\textbf{QWEN/TeleQNA $\rightarrow$ ARC Dataset}} \\
\midrule
Baseline & 0.0268 & 0.0158 & 0.0003 & 0.1997 & 0.8038 & 0.0088 \\
LoRA FineTune & 0.0151 & 0.0101 & 0.0004 & 0.1869 & 0.8080 & 0.0073 \\
LoRA Transfer & 0.0247 & 0.0163 & 0.0006 & 0.2635 & 0.8068 & 0.0069 \\
\midrule
\multicolumn{7}{c}{\textbf{QWEN/TeleQNA $\rightarrow$ MedMCQ Dataset}} \\
\midrule
Baseline & 0.0577 & 0.0179 & 0.0006 & 0.8964 & 0.4805 & 0.0057 \\
LoRA FineTune & 0.0363 & 0.0771 & 0.0082 & 0.9031 & 0.4798 & 0.0055 \\
LoRA Transfer & 0.0484 & 0.0279 & 0.0024 & 0.2639 & 0.4803 & 0.0049 \\
\midrule
\multicolumn{7}{c}{\textbf{QWEN/TeleQNA $\rightarrow$ QASC Dataset}} \\
\midrule
Baseline & 0.0110 & 0.0547 & 0.0117 & 0.2062 & 0.8952 & 0.0068 \\
LoRA FineTune & 0.0031 & 0.0101 & 0.0003 & 0.1188 & 0.9698 & 0.0040 \\
LoRA Transfer & 0.0167 & 0.0913 & 0.0516 & 0.2408 & 0.8337 & 0.0097 \\
\bottomrule
\end{tabular}
\end{table}

\begin{table}[H]
\centering
\scriptsize
\caption{Transferability of Model-Checkpoints (Continued)}
\begin{tabular}{@{}lcccccc@{}}
\toprule
\textbf{Source/Target} & \textbf{PBM} & \textbf{RStd} & \textbf{CKLD} & \textbf{Fluct. Rate} & \textbf{Accuracy} & \textbf{Perm Acc Std} \\
\midrule
\multicolumn{7}{c}{\textbf{QWEN/ARC $\rightarrow$ TeleQNA Dataset}} \\
\midrule
Baseline & 0.0214 & 0.0564 & 0.0070 & 0.5874 & 0.5464 & 0.0189 \\
LoRA FineTune & 0.0182 & 0.0879 & 0.0223 & 0.6093 & 0.5437 & 0.0141 \\
LoRA Transfer & 0.0025 & 0.1363 & 0.0805 & 0.1694 & 0.5464 & 0.0139 \\
\midrule
\multicolumn{7}{c}{\textbf{QWEN/ARC $\rightarrow$ MedMCQ Dataset}} \\
\midrule
Baseline & 0.0577 & 0.0179 & 0.0006 & 0.8964 & 0.4805 & 0.0057 \\
LoRA FineTune & 0.0363 & 0.0771 & 0.0082 & 0.9031 & 0.4798 & 0.0055 \\
LoRA Transfer & 0.0041 & 0.1532 & 0.1005 & 0.4023 & 0.4604 & 0.0079 \\
\midrule
\multicolumn{7}{c}{\textbf{QWEN/ARC $\rightarrow$ QASC Dataset}} \\
\midrule
Baseline & 0.0110 & 0.0547 & 0.0117 & 0.2062 & 0.8952 & 0.0068 \\
LoRA FineTune & 0.0031 & 0.0101 & 0.0003 & 0.1188 & 0.9698 & 0.0040 \\
LoRA Transfer & 0.0214 & 0.2950 & 1.3080 & 0.8293 & 0.3013 & 0.0134 \\
\bottomrule
\end{tabular}
\end{table}

\begin{table}[H]
\centering
\scriptsize
\caption{Transferability of Model-Checkpoints (Continued)}
\begin{tabular}{@{}lcccccc@{}}
\toprule
\textbf{Source/Target} & \textbf{PBM} & \textbf{RStd} & \textbf{CKLD} & \textbf{Fluct. Rate} & \textbf{Accuracy} & \textbf{Perm Acc Std} \\
\midrule
\multicolumn{7}{c}{\textbf{QWEN/QASC $\rightarrow$ ARC Dataset}} \\
\midrule
Baseline & 0.0268 & 0.0158 & 0.0003 & 0.1997 & 0.8038 & 0.0088 \\
LoRA FineTune & 0.0151 & 0.0101 & 0.0004 & 0.1869 & 0.8080 & 0.0073 \\
LoRA Transfer & 0.0279 & 0.0183 & 0.0011 & 0.2481 & 0.7923 & 0.0074 \\
\midrule
\multicolumn{7}{c}{\textbf{QWEN/QASC $\rightarrow$ TeleQNA Dataset}} \\
\midrule
Baseline & 0.0214 & 0.0564 & 0.0070 & 0.5874 & 0.5464 & 0.0189 \\
LoRA FineTune & 0.0182 & 0.0879 & 0.0223 & 0.6093 & 0.5437 & 0.0141 \\
LoRA Transfer & 0.0408 & 0.0620 & 0.0223 & 0.2158 & 0.5601 & 0.0140 \\
\midrule
\multicolumn{7}{c}{\textbf{QWEN/QASC $\rightarrow$ MedMCQ Dataset}} \\
\midrule
Baseline & 0.0577 & 0.0179 & 0.0006 & 0.8964 & 0.4805 & 0.0057 \\
LoRA FineTune & 0.0363 & 0.0771 & 0.0082 & 0.9031 & 0.4798 & 0.0055 \\
LoRA Transfer & 0.0573 & 0.1089 & 0.0067 & 0.2175 & 0.4735 & 0.0044 \\
\bottomrule
\end{tabular}
\end{table}

\begin{table}[H]
\centering
\scriptsize
\caption{Transferability of Phi-2 Model-Checkpoints}
\begin{tabular}{@{}lcccccc@{}}
\toprule
\textbf{Source/Target} & \textbf{PBM} & \textbf{RStd} & \textbf{CKLD} & \textbf{Fluct. Rate} & \textbf{Accuracy} & \textbf{Perm Acc Std} \\
\midrule
\multicolumn{7}{c}{\textbf{Phi-2/TeleQnA $\rightarrow$ ARC Dataset}} \\
\midrule
Baseline & 0.0269 & 0.2207 & 0.2853 & 0.6334 & 0.5520 & 0.0110 \\
LoRA FineTune & 0.0148 & 0.0257 & 0.0021 & 0.2858 & 0.7553 & 0.0076 \\
LoRA Transfer & 0.0217 & 0.0545 & 0.0067 & 0.2576 & 0.6500 & 0.0080 \\
\midrule
\multicolumn{7}{c}{\textbf{Phi-2/TeleQnA $\rightarrow$ MedMCQ Dataset}} \\
\midrule
Baseline & 0.0512 & 0.3981 & 1.4585 & 0.9369 & 0.3409 & 0.0194 \\
LoRA FineTune & 0.0013 & 0.0893 & 0.0527 & 0.3358 & 0.3569 & 0.0087 \\
LoRA Transfer & 0.0256 & 0.0709 & 0.0478 & 0.3799 & 0.3344 & 0.0051 \\
\midrule
\multicolumn{7}{c}{\textbf{Phi-2/TeleQnA $\rightarrow$ QASC Dataset}} \\
\midrule
Baseline & 0.0314 & 0.3232 & 2.9914 & 0.9676 & 0.1577 & 0.0122 \\
LoRA FineTune & 0.0002 & 0.2373 & 1.0080 & 0.1003 & 0.3002 & 0.0112 \\
LoRA Transfer & 0.0073 & 0.0605 & 0.0067 & 0.0950 & 0.8542 & 0.0008 \\
\bottomrule
\end{tabular}
\end{table}

\begin{table}[H]
\centering
\scriptsize
\caption{Transferability of Phi-2 Model-QASC}
\begin{tabular}{@{}lcccccc@{}}
\toprule
\textbf{Source/Target} & \textbf{PBM} & \textbf{RStd} & \textbf{CKLD} & \textbf{Fluct. Rate} & \textbf{Accuracy} & \textbf{Perm Acc Std} \\
\midrule
\multicolumn{7}{c}{\textbf{Phi-2/Qasc $\rightarrow$ ARC Dataset}} \\
\midrule
Baseline & 0.0269 & 0.2207 & 0.2853 & 0.6334 & 0.5520 & 0.0110 \\
LoRA FineTune & 0.0148 & 0.0257 & 0.0021 & 0.2858 & 0.7553 & 0.0076 \\
LoRA Transfer & 0.0129 & 0.2855 & 0.5250 & 0.7511 & 0.4489 & 0.0091 \\
\midrule
\multicolumn{7}{c}{\textbf{Phi-2/Qasc $\rightarrow$ TeleQNA Dataset}} \\
\midrule
Baseline & 0.0303 & 0.3559 & 1.5208 & 0.8114 & 0.2568 & 0.0216 \\
LoRA FineTune & 0.0260 & 0.0271 & 0.0087 & 0.0267 & 0.3530 & 0.0099 \\
LoRA Transfer & 0.0494 & 0.4000 & 2.9890 & 1.0000 & 0.2100 & 0.0233 \\
\midrule
\multicolumn{7}{c}{\textbf{Phi-2/Qasc $\rightarrow$ MedMCQ Dataset}} \\
\midrule
Baseline & 0.0512 & 0.3981 & 1.4585 & 0.9369 & 0.3409 & 0.0194 \\
LoRA FineTune & 0.0013 & 0.0893 & 0.0527 & 0.3358 & 0.3569 & 0.0087 \\
LoRA Transfer & 0.0300 & 0.4259 & 2.4919 & 0.9931 & 0.3201 & 0.0199 \\
\bottomrule
\end{tabular}
\end{table}

\begin{table}[H]
\centering
\scriptsize
\caption{Transferability of Phi-2 Model-MedMCQA)}
\begin{tabular}{@{}lcccccc@{}}
\toprule
\textbf{Source/Target} & \textbf{PBM} & \textbf{RStd} & \textbf{CKLD} & \textbf{Fluct. Rate} & \textbf{Accuracy} & \textbf{Perm Acc Std} \\
\midrule
\multicolumn{7}{c}{\textbf{Phi-2/MedMCQ $\rightarrow$ ARC Dataset}} \\
\midrule
Baseline & 0.0269 & 0.2207 & 0.2853 & 0.6334 & 0.5520 & 0.0110 \\
LoRA FineTune & 0.0148 & 0.0257 & 0.0021 & 0.2858 & 0.7553 & 0.0076 \\
LoRA Transfer & 0.0149 & 0.0257 & 0.0021 & 0.2858 & 0.7553 & 0.0076 \\
\midrule
\multicolumn{7}{c}{\textbf{Phi-2/MedMCQ $\rightarrow$ TeleQNA Dataset}} \\
\midrule
Baseline & 0.0303 & 0.3559 & 1.5208 & 0.8114 & 0.2568 & 0.0216 \\
LoRA FineTune & 0.0260 & 0.0271 & 0.0087 & 0.0267 & 0.3530 & 0.0099 \\
LoRA Transfer & 0.0260 & 0.0960 & 0.0115 & 0.1939 & 0.4125 & 0.0170 \\
\midrule
\multicolumn{7}{c}{\textbf{Phi-2/MedMCQ $\rightarrow$ QASC Dataset}} \\
\midrule
Baseline & 0.0314 & 0.3232 & 2.9914 & 0.9676 & 0.1577 & 0.0122 \\
LoRA FineTune & 0.0002 & 0.2373 & 1.0080 & 0.1003 & 0.3002 & 0.0112 \\
LoRA Transfer & 0.0004 & 0.0291 & 0.0017 & 0.1134 & 0.9265 & 0.0051 \\
\bottomrule
\end{tabular}
\end{table}

\begin{table}[H]
\centering
\scriptsize
\caption{Transferability of Phi-2/ARC Model-ARC}
\begin{tabular}{@{}lcccccc@{}}
\toprule
\textbf{Source/Target} & \textbf{PBM} & \textbf{RStd} & \textbf{CKLD} & \textbf{Fluct. Rate} & \textbf{Accuracy} & \textbf{Perm Acc Std} \\
\midrule
\multicolumn{7}{c}{\textbf{Phi-2/ARC $\rightarrow$ ARC Dataset}} \\
\midrule
Baseline & 0.0269 & 0.2207 & 0.2853 & 0.6334 & 0.5520 & 0.0110 \\
LoRA FineTune & 0.0148 & 0.0257 & 0.0021 & 0.2858 & 0.7553 & 0.0076 \\
LoRA Transfer & 0.0217 & 0.0545 & 0.0067 & 0.2575 & 0.6498 & 0.0076 \\
\midrule
\multicolumn{7}{c}{\textbf{Phi-2/ARC $\rightarrow$ TeleQNA Dataset}} \\
\midrule
Baseline & 0.0303 & 0.3559 & 1.5208 & 0.8114 & 0.2568 & 0.0216 \\
LoRA FineTune & 0.0260 & 0.0271 & 0.0087 & 0.0267 & 0.3530 & 0.0099 \\
LoRA Transfer & 0.0261 & 0.0964 & 0.0115 & 0.1939 & 0.4126 & 0.0099 \\
\midrule
\multicolumn{7}{c}{\textbf{Phi-2/ARC $\rightarrow$ MedMCQ Dataset}} \\
\midrule
Baseline & 0.0512 & 0.3981 & 1.4585 & 0.9369 & 0.3409 & 0.0194 \\
LoRA FineTune & 0.0013 & 0.0893 & 0.0527 & 0.3358 & 0.3569 & 0.0087 \\
LoRA Transfer & 0.0269 & 0.0893 & 0.0214 & 0.2737 & 0.3749 & 0.0045 \\
\multicolumn{7}{c}{\textbf{Phi-2/ARC $\rightarrow$ QASC Dataset}} \\
\midrule
Baseline & 0.0314 & 0.3232 & 2.9914 & 0.9676 & 0.1577 & 0.0122 \\
LoRA FineTune & 0.0002 & 0.2373 & 1.0080 & 0.1003 & 0.3002 & 0.0112 \\
LoRA Transfer & 0.0042 & 0.0291 & 0.0017 & 0.1133 & 0.9265 & 0.0059 \\
\bottomrule
\end{tabular}
\end{table}

\begin{table}[H]
\centering
\scriptsize
\caption{Transferability of Llama 3.2 Model-ARC }
\begin{tabular}{@{}lcccccc@{}}
\toprule
\textbf{Source/Target} & \textbf{PBM} & \textbf{RStd} & \textbf{CKLD} & \textbf{Fluct. Rate} & \textbf{Accuracy} & \textbf{Perm Acc Std} \\
\midrule
\multicolumn{7}{c}{\textbf{Llama 3.2/ARC $\rightarrow$ TeleQNA Dataset}} \\
\midrule
Baseline & 0.0053 & 0.1010 & 0.0907 & 0.5574 & 0.4536 & 0.0165 \\
LoRA FineTune & 0.0002 & 0.1998 & 0.3120 & 0.3306 & 0.3830 & 0.0245 \\
LoRA Transfer & 0.0007 & 0.2344 & 0.5903 & 0.6393 & 0.4235 & 0.0209 \\
\midrule
\multicolumn{7}{c}{\textbf{Llama 3.2/ARC $\rightarrow$ MedMCQ Dataset}} \\
\midrule
Baseline & 0.0167 & 0.3182 & 0.5597 & 0.8128 & 0.3696 & 0.0161 \\
LoRA FineTune & 0.0000 & 0.0052 & 0.0531 & 0.2116 & 0.3213 & 0.0056 \\
LoRA Transfer & 0.0012 & 0.3943 & 1.2546 & 0.9132 & 0.3541 & 0.0189 \\
\midrule
\multicolumn{7}{c}{\textbf{Llama 3.2/ARC $\rightarrow$ QASC Dataset}} \\
\midrule
Baseline & 0.0046 & 0.2021 & 0.4818 & 0.6544 & 0.4892 & 0.0138 \\
LoRA FineTune & 0.0026 & 0.0130 & 0.0014 & 0.1263 & 0.9514 & 0.0037 \\
LoRA Transfer & 0.0020 & 0.2963 & 1.7180 & 0.8920 & 0.2397 & 0.0119 \\
\bottomrule
\end{tabular}
\end{table}

\begin{table}[H]
\centering
\scriptsize
\caption{Transferability of Llama 3.2 Model-TeleQnA}
\begin{tabular}{@{}lcccccc@{}}
\toprule
\textbf{Source/Target} & \textbf{PBM} & \textbf{RStd} & \textbf{CKLD} & \textbf{Fluct. Rate} & \textbf{Accuracy} & \textbf{Perm Acc Std} \\
\midrule
\multicolumn{7}{c}{\textbf{Llama 3.2/TeleQNA $\rightarrow$ ARC Dataset}} \\
\midrule
Baseline & 0.0091 & 0.1082 & 0.1050 & 0.5171 & 0.5179 & 0.0097 \\
LoRA FineTune & 0.0034 & 0.1780 & 0.1144 & 0.1373 & 0.5339 & 0.0092 \\
LoRA Transfer & 0.0010 & 0.1828 & 0.2423 & 0.5416 & 0.4343 & 0.0136 \\
\midrule
\multicolumn{7}{c}{\textbf{Llama 3.2/TeleQNA $\rightarrow$ MedMCQ Dataset}} \\
\midrule
Baseline & 0.0167 & 0.3182 & 0.5597 & 0.8128 & 0.3696 & 0.0161 \\
LoRA FineTune & 0.0000 & 0.0052 & 0.0531 & 0.2116 & 0.3213 & 0.0056 \\
LoRA Transfer & 0.0004 & 0.3840 & 1.6399 & 0.9013 & 0.3376 & 0.0056 \\
\midrule
\multicolumn{7}{c}{\textbf{Llama 3.2/TeleQNA $\rightarrow$ QASC Dataset}} \\
\midrule
Baseline & 0.0046 & 0.2021 & 0.4818 & 0.6544 & 0.4892 & 0.0138 \\
LoRA FineTune & 0.0026 & 0.0130 & 0.0014 & 0.1263 & 0.9514 & 0.0037 \\
LoRA Transfer & 0.0012 & 0.3290 & 3.5746 & 0.9946 & 0.1369 & 0.0124 \\
\bottomrule
\end{tabular}
\end{table}

\begin{table}[H]
\centering
\scriptsize
\caption{Transferability of Llama 3.2 Model-QASC}
\begin{tabular}{@{}lcccccc@{}}
\toprule
\textbf{Source/Target} & \textbf{PBM} & \textbf{RStd} & \textbf{CKLD} & \textbf{Fluct. Rate} & \textbf{Accuracy} & \textbf{Perm Acc Std} \\
\midrule
\multicolumn{7}{c}{\textbf{Llama 3.2/QASC $\rightarrow$ ARC Dataset}} \\
\midrule
Baseline & 0.0091 & 0.1082 & 0.1050 & 0.5171 & 0.5179 & 0.0097 \\
LoRA FineTune & 0.0034 & 0.1780 & 0.1144 & 0.1373 & 0.5339 & 0.0092 \\
LoRA Transfer & 0.0003 & 0.1195 & 0.1184 & 0.4515 & 0.4944 & 0.0094 \\
\midrule
\multicolumn{7}{c}{\textbf{Llama 3.2/QASC $\rightarrow$ TeleQNA Dataset}} \\
\midrule
Baseline & 0.0053 & 0.1010 & 0.0907 & 0.5574 & 0.4536 & 0.0165 \\
LoRA FineTune & 0.0002 & 0.1998 & 0.3120 & 0.3306 & 0.3830 & 0.0245 \\
LoRA Transfer & 0.0007 & 0.2344 & 0.5903 & 0.6393 & 0.4235 & 0.0175 \\
\midrule
\multicolumn{7}{c}{\textbf{Llama 3.2/QASC $\rightarrow$ MedMCQ Dataset}} \\
\midrule
Baseline & 0.0167 & 0.3182 & 0.5597 & 0.8128 & 0.3696 & 0.0161 \\
LoRA FineTune & 0.0000 & 0.0052 & 0.0531 & 0.2116 & 0.3213 & 0.0056 \\
LoRA Transfer & 0.0012 & 0.3943 & 1.2545 & 0.9132 & 0.3541 & 0.0018 \\
\bottomrule
\end{tabular}
\end{table}

\begin{table}[H]
\centering
\scriptsize
\caption{Transferability of Llama 3.2 Model-MedMCQ}
\begin{tabular}{@{}lcccccc@{}}
\toprule
\textbf{Source/Target} & \textbf{PBM} & \textbf{RStd} & \textbf{CKLD} & \textbf{Fluct. Rate} & \textbf{Accuracy} & \textbf{Perm Acc Std} \\
\midrule
\multicolumn{7}{c}{\textbf{Llama 3.2/MedMCQ $\rightarrow$ ARC Dataset}} \\
\midrule
Baseline & 0.0091 & 0.1082 & 0.1050 & 0.5171 & 0.5179 & 0.0097 \\
LoRA FineTune & 0.0034 & 0.1780 & 0.1144 & 0.1373 & 0.5339 & 0.0092 \\
LoRA Transfer & 0.0011 & 0.0914 & 0.0645 & 0.4129 & 0.4893 & 0.0133 \\
\midrule
\multicolumn{7}{c}{\textbf{Llama 3.2/MedMCQ $\rightarrow$ TeleQNA Dataset}} \\
\midrule
Baseline & 0.0053 & 0.1010 & 0.0907 & 0.5574 & 0.4536 & 0.0165 \\
LoRA FineTune & 0.0002 & 0.1998 & 0.3120 & 0.3306 & 0.3830 & 0.0245 \\
LoRA Transfer & 0.0003 & 0.1730 & 0.2920 & 0.4508 & 0.4153 & 0.0202 \\
\midrule
\multicolumn{7}{c}{\textbf{Llama 3.2/MedMCQ $\rightarrow$ QASC Dataset}} \\
\midrule
Baseline & 0.0046 & 0.2021 & 0.4818 & 0.6544 & 0.4892 & 0.0138 \\
LoRA FineTune & 0.0026 & 0.0130 & 0.0014 & 0.1263 & 0.9514 & 0.0037 \\
LoRA Transfer & 0.0009 & 0.2999 & 1.8713 & 0.9039 & 0.2441 & 0.0154 \\
\bottomrule
\end{tabular}
\end{table}


\section{Prompt Templates and Hyperparameter Configurations}\label{appendix:prompt-template}
\subsection{Prompt Templates Used in All Experiments}

This appendix provides the exact prompt templates used during evaluation of transferability experiments across datasets and models.

\subsection*{MedMCQA Dataset (All Models)}
\begin{verbatim}
Instruct = Youre a Medical Question Answering Expert, answer the following question. 
Please generate only answer choice (1, 2, 3 or 4)
{question}
{options}
Output: option 
\end{verbatim}

\subsection*{TeleQnA Dataset (Qwen \& Llama)}
\begin{verbatim}
Instruct: Answer the following question using the context provided.
Your answer must start with the correct option letter (1, 2, 3, 4, or 5):
{question}
{options}
Output: option 
\end{verbatim}

\subsection*{TeleQnA or ARC Dataset (Phi-2)}
\begin{verbatim}
Instruct: Answer the following question.
Your answer must start with the correct option letter (1, 2, 3, 4, or 5) followed by the 
text of the answer:
{question}
{options}
Output: option 
\end{verbatim}

\subsection*{ARC Dataset (All Models Except Phi)}
\begin{verbatim}
Instruct: {question}
{options}
Output: option 
\end{verbatim}

\subsection*{QASC Dataset (Qwen, Phi-2)}
\begin{verbatim}
Instruct: Answer the following question using the context provided, reason over it. 
Please generate only answer choice (1, 2, 3, 4, 5, 6, 7 or 8) without any explanations

{question}
context: {context}      

{options}
{question}
Output: option 
\end{verbatim}

\subsection*{QASC Dataset (Llama)}
\begin{verbatim}
Instruct: Answer the following question using the context provided, reason over it. 
Please generate only answer choice (1, 2, 3, 4, 5, 6, 7 or 8) without any explanations

{question}
context: {context}

{options}
Output: 
\end{verbatim}

\subsection{Hyper-Parameter Configurations}
\textbf{Unsupervised LoRA Finetuning:} During each iteration we randomly sample 64 samples every epoch from the train split of the dataset and use it for training. For the LoRA adapters, we target the attention  QKV and Output projection weights, with a dropout of 0.05 and LoRA alpha of 16. We used an AdamW optimizer with learning rate of $1e^{-4}$ and weight decay of $0.001$ for all experiments. In computing the loss, we find the use of $\alpha$ of 0.1 to weight the entropy as yielding good performance. However, further experiments on the balance of this hyper-parameter choice can be further studied in more details. 
\subsection{Model Size And Budget}
We evaluated our methods on a set of compact yet capable LLMs, including Qwen2.5–3B, Phi-2, and LLaMA3.2–3B, each with approximately 3 billion parameters. All experiments were conducted on NVIDIA L40 GPUs (48 GB VRAM).
Using AWS g5.12xlarge instances (approximate L40 equivalent) priced at \$3.06/hour on-demand, the cost per fine-tuning run is \$6–\$9, and BaQCKV inference costs are \$1–\$2 per dataset. This keeps the total cost for running all experiments within a practical research budget, demonstrating that our methods are efficient and deployable even on mid-sized mod.\\
\textbf{Bias Metric Computation}: To compute the PBM, we do not use the raw probabilities over the vocab size. We rather use the logits for only the option IDs to compute the probabilities. Also, we limit the number of permutations to a maximum of 24 ($4!$). This is due to our GPU memory budget and our considerations that most MCQs would have around 4 options.
\subsection{Dataset Description}\label{appendix:dataset-description}

We evaluate our methods on four publicly‑available multiple‑choice QA benchmarks: TeleQnA \citep{maatouk2023teleqnabenchmarkdatasetassess}, MedMCQA\citep{pal2022medmcqalargescalemultisubject}, QASC\citep{khot2020qascdatasetquestionanswering} and ARC Challenge\citep{Clark2018ThinkYH}. Table~\ref{tab:dataset_stats} summarizes their key statistics.

\begin{table}[ht]
  \centering
  \small
  \setlength{\tabcolsep}{4pt}
  \renewcommand{\arraystretch}{1.1}
  \begin{tabular}{@{}lcccc@{}}
    \toprule
    \textbf{Dataset}     & \textbf{Domain}  &  \textbf{\# Choices} & \textbf{Splits (Train/Test)} & \textbf{License}      \\
    \midrule
    TeleQnA             & Telecom knowledge             &   2--5                 & 1461 / 366 & MIT                  \\
    MedMCQA             & Medical exam    &  4    & 5000/ 4183 & MIT                  \\
    QASC                & Grade‑school science  &  8                    & 8134 /926         & CC BY 4.0            \\
    ARC Challenge       & Grade‑school science (hard)    & 4                    & 1119 /1172      & CC BY‑SA 4.0         \\
    \bottomrule
  \end{tabular}
  \caption{Overview of QA benchmarks used in our experiments.}
  \label{tab:dataset_stats}
\end{table}

\textbf{TeleQnA} is a telecommunications‑domain multiple‑choice QA dataset consisting of 10,000 English questions extracted from 3GPP standards and research articles. Each question has between 2 and 5 answer options. We use the publicly released splits for the Zindi subset of the data \mbox{\cite{Zindi}}. TeleQnA is distributed under the MIT license. 
\\
\textbf{MedMCQA} comprises 194,000 medical‑entrance exam questions across 21 subjects and over 2,400 topics. Each item has 4 answer choices. We used the validation set as our test set in order to have access to the answers. It is available under the MIT license.
\\
\textbf{QASC} is a multi‑hop commonsense reasoning dataset containing  9,980 grade‑school science questions designed for sentence composition. Each question has 8 choices. We follow the official splits of 8,134 training, 926 validation, and 920 test examples. We also used the validation set as our test, similar to the MedMCQA. It is released under a CC BY 4.0 license. 
\\
\textbf{The ARC Challenge} set contains 2,590 ``hard'' grade‑school science questions that simple IR baselines fail to answer correctly. All questions have 4 answer options. We adopt the official splits: 1,119 train and 1,172 test examples. We use the actual test split as our test split since it contains the answers. The dataset is provided under CC BY‑SA 4.0.

\section{Computational Efficiency Comparison of the Bias Mitigation Strategies} \label{appendix:computational-efficiency}


\begin{table}[h]
\centering
\scriptsize
\setlength{\tabcolsep}{4pt}
\begin{tabular}{lrrrrrrr}
\toprule
\multirow{2}{*}{Mitigation Method} & \multicolumn{2}{c}{ARC Dataset} & \multicolumn{2}{c}{MEDMCQ Dataset} & \multicolumn{2}{c}{QASC Dataset} \\
\cmidrule(lr){2-3} \cmidrule(lr){4-5} \cmidrule(lr){6-7}
 & Debiasing Time (s) & Inference Time (s) & Debiasing Time (s) & Inference Time (s) & Debiasing Time (s) & Inference Time (s) \\
\midrule
LoRA-1   & 143.30 & 0.042  & 91.33   & 0.059  & 100.03 & 0.058 \\
Gray     & -      & 0.049  & -       & 0.061  & -      & 0.0796 \\
BNP      & 182.87 & 0.037  & 74.87   & 0.036  & 242.48 & 0.036 \\
PriDe    & 52.54  & 0.042  & 248.27  & 0.037  & 87.20  & 0.037 \\
\bottomrule
\end{tabular}
\vspace{-0.1in}
\caption{
Comparison of the computational time for the different mitigation methods on the ARC, MEDMCQ, and QASC datasets. The debiasing time refers to the time required to compute calibration parameters or fine-tune the model before inference. The average inference time indicates the time to compute the final answer logits per example.
}
\label{tab:mitigation_times}
\end{table}

From \mbox{Table~\ref{tab:mitigation_times}}, it can be seen that BNP consistently incurs the longest debiasing time across datasets. This is due to the overhead of computing the bias vector, which in the worst case requires a forward pass over all permutations of the dataset. In contrast, our proposed method, LoRA-1, offers a more efficient trade-off between debiasing time and inference speed. Specifically, LoRA-1 finetuning takes an average of 28.6 seconds per epoch using the BaQKCV loss function, resulting in approximately 143 seconds for five epochs on the ARC dataset. For MEDMCQ and QASC, LoRA-1 required only 91.3s and 100.0s, respectively, which are both lower than BNP and PriDe in those settings.While PriDe required just 52.5s on ARC, its debiasing time rose significantly on the MEDMCQ dataset (248.3s), highlighting its sensitivity to dataset size as it calibrates using 10\% of the data, and so its cost scales with dataset volume. On the QASC dataset, its time (87.2s) was also slightly lower than LoRA-1. However, as dataset sizes grow, LoRA-1 remains more stable and predictable, making it more suitable for scalable deployment. Additionally, LoRA-1 maintains competitive inference times, showing only minor overhead compared to GrayBox and BNP.


\end{document}